\documentclass[review,10pt,authoryear]{elsarticle}

\usepackage{graphicx}
\usepackage{url}
\usepackage{enumitem}
\usepackage{amsfonts}
\usepackage{amsmath}
\usepackage{amssymb}
\usepackage{booktabs}
\usepackage{enumitem}
\usepackage{boldline}
\usepackage{threeparttable}
\usepackage{array}
\usepackage{float}
\usepackage{textcomp}
\usepackage{mdwmath}
\usepackage{mdwtab}
\usepackage{subfig}
\usepackage{amsmath}
\usepackage{color}
\usepackage{tikz}
\usepackage{multirow}
\usepackage{bbding}
\usepackage{caption}
\usepackage{bm}
\usepackage{amsthm}
\usepackage{dblfloatfix}
\usepackage{hyperref}
\usepackage{algorithm,algpseudocode}
\usepackage{lipsum}
\usepackage{etoolbox}
\usepackage{lineno}
\usepackage{tikz}
\usepackage[margin=2cm]{geometry}
\usepackage{pdflscape}

\definecolor{navy}{RGB}{0, 0, 128}

\makeatletter
\newenvironment{breakablealgorithm}
  {
   \begin{center}
     \refstepcounter{algorithm}
     \hrule height.8pt depth0pt \kern2pt
     \renewcommand{\caption}[2][\relax]{
       {\raggedright\textbf{\ALG@name~\thealgorithm:} ##2\par}%
       \ifx\relax##1\relax 
         \addcontentsline{loa}{algorithm}{\protect\numberline{\thealgorithm}##2}%
       \else 
         \addcontentsline{loa}{algorithm}{\protect\numberline{\thealgorithm}##1}%
       \fi
       \kern2pt\hrule\kern2pt
     }
  }{
     \kern2pt\hrule\relax
   \end{center}
  }
\makeatother

\algnewcommand\algorithmicinput{\textbf{Input:}}
\algnewcommand\Input{\item[\algorithmicinput]}
\algnewcommand\algorithmicoutput{\textbf{Output:}}
\algnewcommand\Output{\item[\algorithmicoutput]}
\algnewcommand\algorithmicParameter{\textbf{Parameters:}}
\algnewcommand\Parameters{\item[\algorithmicParameter]}

\journal{Expert Systems with Applications}

\begin{document}

\begin{frontmatter}

\captionsetup[figure]{labelfont={bf},labelformat={default},labelsep=period,name={Fig.}}
\captionsetup[table]{labelfont={bf},labelformat={default},labelsep=period,name={Table}}

\title{A Modular, Data-Free Pipeline for Multi-Label Intention Recognition in Transportation Agentic AI Applications}


\author[mymainaddress]{Xiaocai Zhang}

\author[mysecondaryaddress]{Hur Lim}

\author[mythirdaddress]{Ke Wang\corref{mycorrespondingauthor}}
\cortext[mycorrespondingauthor]{Corresponding author.}
\ead{wang\_ke@ihpc.a-star.edu.sg}

\author[mythirdaddress]{Zhe Xiao}

\author[myfourthaddress]{Jing Wang}

\author[mythirdaddress]{Kelvin Lee}

\author[mythirdaddress]{Xiuju Fu}

\author[mythirdaddress]{Zheng Qin}

\address[mymainaddress]{Department of Infrastructure Engineering, Faculty of Engineering and Information Technology, The University of Melbourne, Australia}
\address[mysecondaryaddress]{School of Computing, National University of Singapore, Singapore}
\address[mythirdaddress]{Institute of High Performance Computing, Agency for Science, Technology and Research (A*STAR), Singapore}
\address[myfourthaddress]{Department of Civil and Environmental Engineering, The Hong Kong University of Science and Technology, Hong Kong SAR}

\begin{abstract}
In this study, a modular, data-free pipeline for multi-label intention recognition is proposed for agentic AI applications in transportation. Unlike traditional intent recognition systems that depend on large, annotated corpora and often struggle with fine-grained, multi-label discrimination, our approach eliminates the need for costly data collection while enhancing the accuracy of multi-label intention understanding. Specifically, the overall pipeline, named DMTC, consists of three steps: 1) using prompt engineering to guide large language models (LLMs) to generate diverse synthetic queries in different transport scenarios; 2) encoding each textual query with a Sentence-T5 model to obtain compact semantic embeddings; 3) training a lightweight classifier using a novel online focal-contrastive (OFC) loss that emphasizes hard samples and maximizes inter-class separability. The applicability of the proposed pipeline is demonstrated in an agentic AI application in the maritime transportation context. Extensive experiments show that DMTC achieves a Hamming loss of 5.35\% and an AUC of 95.92\%, outperforming state-of-the-art multi-label classifiers and recent end-to-end SOTA LLM-based baselines. Further analysis reveals that Sentence-T5 embeddings improve subset accuracy by at least 3.29\% over alternative encoders, and integrating the OFC loss yields an additional 0.98\% gain compared to standard contrastive objectives. In conclusion, our system seamlessly routes user queries to task-specific modules (e.g., ETA information, traffic risk evaluation, and other typical scenarios in the transportation domain), laying the groundwork for fully autonomous, intention-aware agents without costly manual labelling.
\end{abstract}


\end{frontmatter}

\section{Introduction}
Over the past decades, Artificial Intelligence (AI) has transformed transportation, empowering applications such as estimated time of arrival (ETA) prediction \citep{zhang2023prediction}, fuel consumption estimation \citep{chen2023prediction}, traffic hotspot detection \citep{xiao2022next}, vessel trajectory forecasting \citep{zhang2024dynamic,zhang2022vessel,liu2023multi}, and vessel type recognition \citep{zhang2024viewpoint} etc. Building on advances in large language models (LLMs) and reinforcement learning, agentic AI—autonomous systems that perceive, plan, and act with minimal oversight—is poised to deliver even greater autonomy in routing, scheduling, and decision‑making in transportation sector. 
However, accurate multi‑label intention understanding remains a bottleneck: mapping free‑form user queries to precise operational goals requires fine‑grained discrimination across high‑dimensional outputs, label correlations, and imbalanced classes. Besides, collecting large, annotated datasets in the transportation domain is labor‑intensive, and existing classifiers often struggle with hard samples.
To address the challenges of costly data collection and low accuracy in multi-label intention understanding within agentic transportation systems, we introduce \textbf{DMTC} (Data-less Multi-label Text Classification), a modular, data-free framework designed to minimize data requirements while achieving high accuracy in multi-label intention understanding. The key contributions of this work are: 
\begin{itemize}
    \item \textbf{Modular pipeline for multi-label intention recognition}: A scalable framework that facilitates seamless integration with agentic AI systems in diverse transportation domains.
    \item \textbf{Zero-shot synthetic data generation via prompt engineering}: An approach that eliminates the need for manual data annotation by generating synthetic datasets, providing valuable benchmarks for the transportation AI community.
    \item \textbf{Novel online focal-contrastive loss function}: A novel loss function that enhances model performance in multi-label classification tasks by focusing on hard-to-classify samples and improving feature discrimination. 
\end{itemize}

\section{RELATED WORK}

Intention-based text classification has been a core task in natural language processing, underpinning applications in domains as diverse as healthcare, finance, and transportation. Advances in agentic AI—autonomous systems that perceive, plan, and act with minimal oversight—have further extended this functionality, enabling automated model orchestration through precise intent detection from user textual queries. 

Text classification methods in transportation typically fall into five categories: rule‑based approaches, traditional machine learning (excluding neural networks and LLMs in this study), neural network models, large language models (LLMs), and hybrid systems \citep{zhang2025natural}. 

Rule‑based methods apply predefined linguistic rules or keyword dictionaries to interpret text. In the study by \cite{maghrebi2015complementing}, they constructed a tweet‑based activity classifier using a manually curated keyword list to label posts into seven travel‑related categories \citep{maghrebi2015complementing}. 

Traditional machine learning methods in text classification predominantly employ SVM and Naive Bayes classifiers. In the work by \cite{ali2019fuzzy}, they trained an SVM to distinguish transport‑related from non‑transportation texts. \cite{styawati2022sentiment} generated Word2Vec embeddings for Gojek and Grab user reviews before applying SVM for sentiment analysis. \cite{klaithin2016traffic} used Naive Bayes to categorize Twitter comments into six traffic‑event classes (e.g., accidents, announcements). \cite{suat2022extraction} combined Doc2Vec embeddings with SVM to classify tweets as accident‑related or not. \cite{teske2018automatic} employed a Bayesian network to identify maritime incident descriptions in news articles.

Advanced neural network methods—particularly RNNs (LSTM/GRU) and CNNs—have become the dominant approach for transportation‑related text classification. \citep{chen2018detecting}  combined CBOW embeddings with an LSTM–CNN hybrid to distinguish traffic‑relevant from traffic‑irrelevant content.   \cite{ali2019fuzzy} employed Bi‑LSTM for sentiment analysis on social media posts and ITS office reports. \cite{chen2017convolutional} used CBOW embeddings with CNN to detect traffic‑related microblogs, outperforming both SVM and MLP baselines. \cite{azhar2023detection} evaluated GloVe embeddings with RNN, GRU, and LSTM for traffic‑tweet detection and severity classification, with LSTM achieving the highest accuracy (94.2 \%). \cite{raksachat2023improving} applied byte‑pair encoding for data augmentation followed by a CNN–Bi‑LSTM pipeline to classify tweets into traffic, accident, disaster, and social‑event categories. \cite{pan2023ernie} introduced an Ernie‑Gram–Bi‑GRU attention architecture for air‑traffic‑control instruction recognition.

Among LLM-based methods in transportation, BERT and its variants have become the de facto standard. \cite{wan2020empowering} applied the bidirectional encoder representations from BERT to classify tweets as traffic‑related or unrelated. \cite{osorio2021social} leveraged BERT for sentiment analysis and multi‑class classification of Madrid Metro tweets—covering incidents, road closures, construction, delays, public transport, and other information—achieving 99.37 \% accuracy against NB, decision tree, and SVM baselines. \cite{zuluaga2022atco2} explored BERT for named‑entity recognition in ATC commands and for detecting speaker roles (e.g., controller vs. pilot) in dialog. \cite{babbar2023real} demonstrated that RoBERTa—an optimized BERT pretraining—outperformed Word2Vec, GloVe, FastText, BERT, and XLNet on sentiment classification, reaching 97 \% accuracy . \cite{zhang2024maritime} introduced a focal‑loss‑trained ConvBERT model for maritime versus non‑maritime text classification, achieving an F1 score of 99.97 \%. Finally, \cite{pullanikkat2024utilizing} evaluated multiple transformer architectures (GPT, BERT, RoBERTa, BERTweet, CrisisTransformer) for public sentiment analysis of transportation issues on Twitter. 
Hybrid methods combine multiple techniques to leverage their complementary strengths. \cite{jidkov2020enabling} used BERT for contextual embeddings, then applied ANN, CNN, and LSTM architectures to classify maritime‑incident texts.  \cite{neruda2021traffic} similarly extracted BERT features from tweets and employed a CNN to detect traffic events, reporting that BERT outperformed ELMo and Word2Vec, with LSTM achieving 94.4\% accuracy. \cite{lee2024analysis} semantically tagged NAVTEX safety messages using a Bi‑LSTM+CRF pipeline aligned with the Common Maritime Data Structure. \cite{khodadadi2022natural} combined TF‑IDF, POS tagging, and n‑grams with $\chi^2$ feature selection, then used a hybrid LSTM‑CNN model to classify customer service claims—routing valid requests via gradient tree boosting. \cite{hui2023atcbert} fine‑tuned BERT on ATC communications with additional masked‑language model training, enabling few‑shot intent recognition in air‑traffic control dialogues.

While rule‑based, traditional ML, neural networks, LLMs, and hybrid approaches each offer valuable capabilities for transportation text classification, none simultaneously eliminate the need for annotated data, support fine‑grained, multi‑label intent detection, and integrate seamlessly into agentic AI workflows. To fill this gap, a modular, data‑free pipeline is designed specifically for zero‑annotation, multi‑label intention recognition in autonomous transportation systems.

\section{METHOD}
\begin{figure*}[htbp]
\begin{center}
\includegraphics[width=0.99\textwidth]{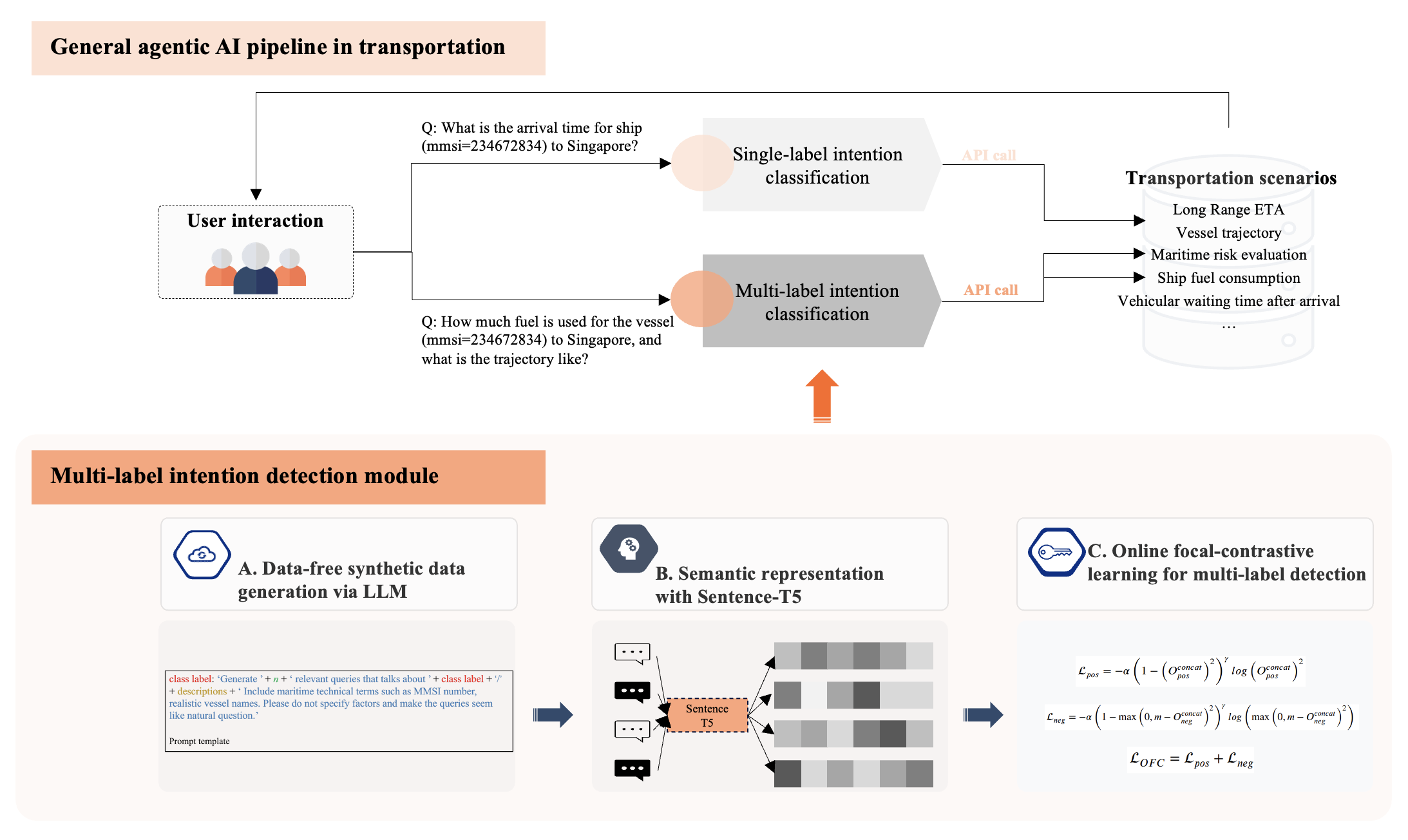}
\caption{The proposed DMTC pipeline for multi-label intention recognization in the maritime transportation}
\label{workflow}
\end{center}
\end{figure*}

Figure. \ref{workflow} outlines the workflow of DMTC for data-free multi-label intention understanding, as implemented in a maritime context. The pipeline comprises three core components: (A) Synthetic data generation via prompt-engineered LLMs; (B) Semantic representation with sentence-T5; (C) Online focal-contrastive learning for multi-label classification.

\subsection{Data-free synthetic data generation with LLM}
\begin{figure}[htbp]
\begin{center}
\includegraphics[width=0.65\linewidth, keepaspectratio]{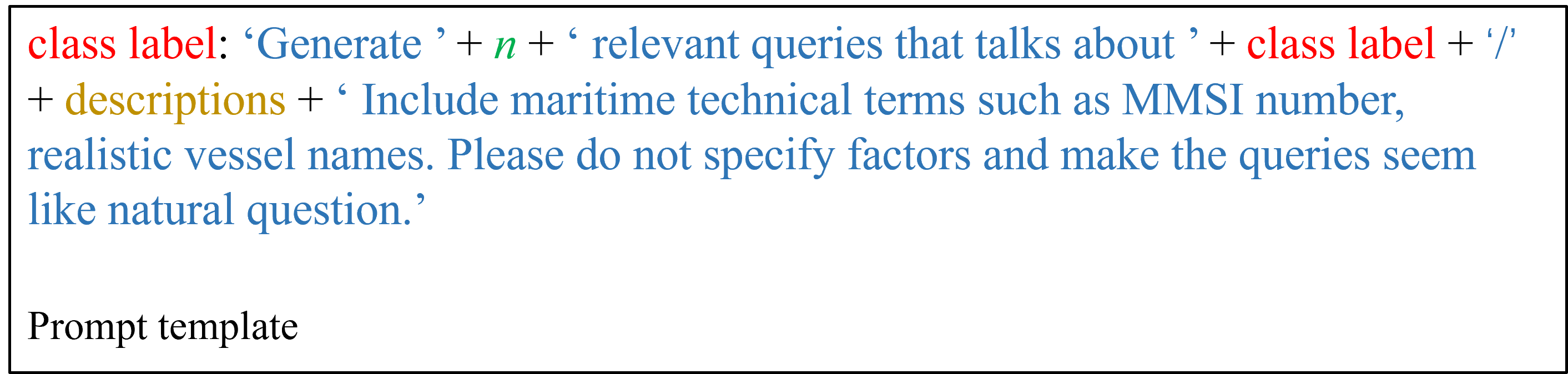}
\caption{Template of prompt for generating data for each class.}
\label{prompt}
\end{center}
\end{figure}

Prompt engineering is employed to generate synthetic datasets without relying on pre-existing data. As illustrated in Figure \ref{prompt}, a unified prompt template is used for this data generation process. The prompt includes the class label, its description, and the desired sample size, where the 'description' provides a clear and concise sentence elaborating on the characteristics of the specified class. 

\begin{figure}[htbp]
\begin{center}
\includegraphics[width=0.65\linewidth, keepaspectratio]{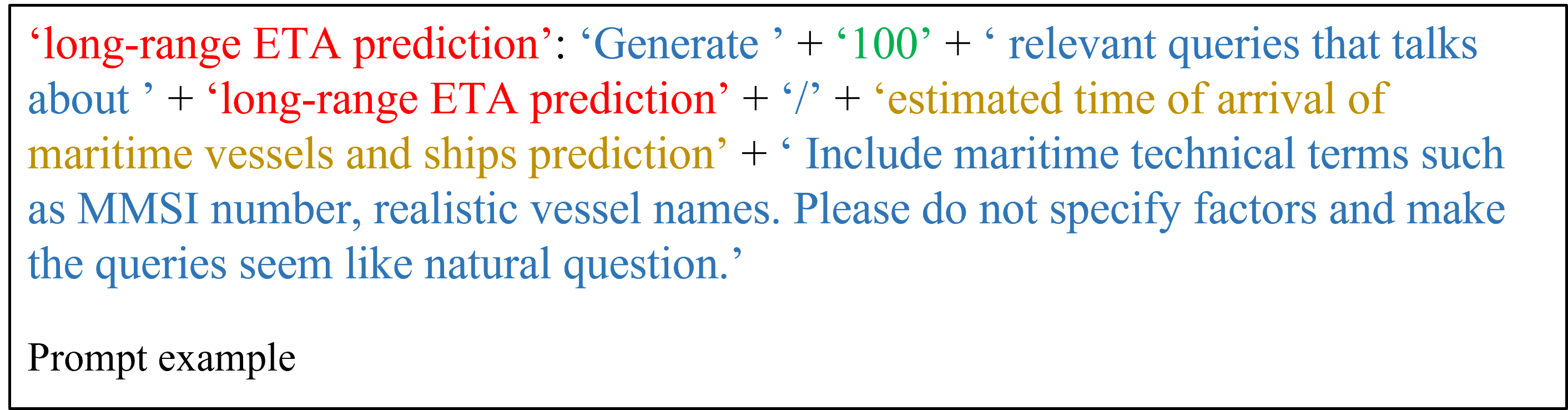}
\caption{An example of constructing the prompt for generating data.}
\label{prompt_example}
\end{center}
\end{figure}

Figure. \ref{prompt_example}  illustrates the prompt for the “long‑range ETA in maritime” class. Here, the class label is set to "long‑range ETA in maritime" and $n$ is set to 100, targeting the generation of 100 samples. The prompt also includes the description “estimated time of arrival for maritime vessels and ships” to clarify the scenario. Once each class template is defined, its prompt is submitted to LLaMA 2 to automate synthetic data creation. For multi-label sample generation, we concatenate the individual text segments corresponding to each label. This plug‑and‑play process can be readily applied to other agentic AI systems.
\subsection{Semantic representation with Sentence-T5}
Sentence-T5 \citep{ni2021sentence}, a derivative of the T5 model, is implemented to learn the contextual representations of texts. It consists of an encoder $E$, and a decoder $D$, both built upon the Transformer architecture. In the pre-training stage, the outputs of the Sentence-T5 are formulated as
\begin{equation}\label{eq1}
O=D\left ( \textup{Pool}\left ( E\left ( S;\theta _{E} \right ) \right );\theta _{D} \right ),
\end{equation}
Where $S$ is the input text, $\theta _{E}$ and $\theta _{D}$ are the parameters of the encoder and decoder, respectively, and Pool represents the average pooling operation.
The encoder $E$ transforms the input text into a sequence of contextual embedding blocks, as represented by
\begin{equation}\label{eq2}
E\left ( S;\theta _{E} \right ) = E_{N}\left ( \cdots E_{1}\left ( S;\theta _{E_{1}} \right );\theta _{E_{N}} \right ),
\end{equation}
where $\theta _{E_{i}}$ denotes the parameters of the $i$th layer of the Transformer.
The average pooling operation is applied across the sequence of encoder outputs to create a pooled representation 
$H$, which is a distinct attribute of the Sentence-T5 to T5 model. The process is formulated as
\begin{equation}\label{eq3}
H = \textup{Pool}\left ( E\left ( S;\theta _{E} \right ) \right ).
\end{equation}
The decoder $D$ uses the pooled representation $H$ to generate the output, as described by
\begin{equation}\label{eq4}
O = D\left ( H;\theta _{D} \right ).
\end{equation}

To retain the sequential structure of the input text, positional encodings (PE) are incorporated into the embeddings, ensuring the model accounts for word order, as shown by
\begin{equation}\label{eq6}
E_{1}\left ( S \right )=E_{word}\left ( S \right )+PE,
\end{equation}
where $PE$ represents the positional encoding vector. 

\subsection{Online focal-contrastive (OFC) learning for multi-label classification}
A novel contrastive loss function, termed OFC loss, is introduced for multi-label intention classification. Designed to capture distinctive features by emphasizing hard samples, OFC incorporates hard‑negative sampling strategies to strengthen inter‑class separability and handle complex label combinations. 
The initial step in OFC involves the creation of pairs comprising positive samples ($S_{pos}$), representing instances with shared labels within the pair, and negative samples ($S_{neg}$), consisting of samples with disparate labels,
\begin{equation}\label{pos}
S_{pos}=\left \{ \left (a_{i}^{l_{a}},b_{i}^{l_{b}}  \right ) |\ i=1,2,\cdots ,n  \right \},
\end{equation}
where $l_{a}=l_{b}$ in $S_{pos}$ and $l_{c}\neq l_{d}$ in $S_{neg}$. The similarity between two instances, i.e., $a^{l_{a}}$ and $b^{l_{b}}$, is defined as
\begin{equation}\label{dist}
Sim\left ( a^{l_{a}},b^{l_{b}} \right )=\frac{a^{l_{a}}\cdot b^{l_{b}}}{\left \| a^{l_{a}} \right \|\cdot \left \| b^{l_{b}} \right \|}.
\end{equation}
Subsequently, we compute the similarity for each instance pair in $S_{pos}$ and $S_{neg}$, for example,
\begin{equation}\label{dist_pos}
D_{pos}=\left \{ Sim\left ( a_{i}^{l_{a}},b_{i}^{l_{b}} \right ) |\ i=1,2,\cdots ,n  \right \},
\end{equation}
Next, hard sample pairs—instances that are challenging to classify correctly—are introduced. Specifically, for positive sample pairs, hard sample pairs is defined using the following formulations, as elucidated by
{\small
\begin{equation}\label{eq7}
H_{pos}=\left \{ \left (a_{i}^{l_{a}},b_{i}^{l_{b}}  \right ), \ \textup{if} \ Sim\left ( a_{i}^{l_{a}},b_{i}^{l_{b}} \right )>T_{neg} \ |\ i=1,2,\cdots ,n  \right \}
\end{equation}
}
and
\begin{equation}\label{eq8}
T_{neg}=\textup{min}\left ( D_{neg} \right),
\end{equation}
where $T_{neg}$ is a threshold determined by the minimum similarity within the similarity set of the negative samples.
Similarly, the identification of hard sample pairs from negative samples $H_{neg}$ follows an opposite process.
Furthermore, the refined positive samples $O_{pos}$ are derived by calculating the relative complement of $H_{pos}$ within $S_{pos}$,
\begin{equation}\label{eq11}
O_{pos}=S_{pos} - H_{pos}.
\end{equation}
Similarly, the refined negative samples $O_{neg}$ are obtained by 
\begin{equation}\label{eq12}
 O_{neg}=S_{neg} - H_{neg}.
 \end{equation}
Following that, we re-arrange the refined positive samples $O_{pos}$ in descending order, as demonstrated by
{\small
\begin{equation}\label{eq13}
O_{pos}^{sorted}= \left \{ Sim\left (  a_{(1)}^{l_{a}},b_{(1)}^{l_{b}} \right ),Sim\left (  a_{(2)}^{l_{a}},b_{(2)}^{l_{b}} \right ),\cdots ,Sim\left (  a_{(k)}^{l_{a}},b_{(k)}^{l_{b}} \right ) \right \}.
\end{equation}
}
On the contrary, we organize the refined positive samples $O_{neg}$ in ascending order.
We choose the initial $p$ percent from the sorted positive and negative samples to create $O_{pos}^{top}$ and $O_{neg}^{top}$, as defined by
\begin{equation}\label{eq17}
O_{pos}^{top}=\textup{select\_top\_element}\left ( O_{pos}^{sorted},p \right ),
\end{equation}
where $p$ is a percentage ratio and $\textup{select\_top\_element}$ is utilized to extract the initial $p\%$ of the whole elements. Subsequently, we combine these initially selected $p\%$ elements with the extracted hard sample pairs to form both the positive samples $O_{pos}^{concat}$ and the negative samples $O_{neg}^{concat}$,
\begin{equation}\label{eq18_1}
O_{pos}^{concat}=\textup{concat}\left ( O_{pos}^{sorted}, H_{pos} \right ),
\end{equation}
The loss for the positive samples is calculated by
\begin{equation}\label{eq19}
\mathcal{L}_{pos}= -\alpha \left ( 1-\left ( O_{pos}^{concat} \right )^{2} \right )^{\gamma }log\left ( O_{pos}^{concat} \right )^{2}.
\end{equation}
Meanwhile, the loss for the negative samples is determined by 
{\footnotesize
\begin{equation}\label{eq20}
\mathcal{L}_{neg}= -\alpha \left ( 1-\textup{max}\left (0,m-O_{neg}^{concat}  \right )^{2} \right )^{\gamma }log\left ( \textup{max}\left (0,m-O_{neg}^{concat}  \right )^{2} \right )
\end{equation}
}
where $m$ is a positive margin parameter and $m>0$. The parameters $\alpha$ and $\gamma$ refer to the weight factor and the focusing parameter, respectively.
Following that, we define the OFC loss as the sum of the positive loss and the negative loss,
\begin{equation}\label{eq21}
\mathcal{L}_{OFC}= \mathcal{L}_{pos}+\mathcal{L}_{neg}.
\end{equation}

The overall learning process can be described as below:  Given a sequence of tokens denoted as a series of tuples, as shown by 
\begin{equation}\label{inp}
X=\left \{x_{t} \ |\ t =1,2,\cdots ,n\right \},
\end{equation}
this sequence $X$ is fed into the pre-trained Sentence-T5 model to generate its corresponding representation.
This resulting representation is expressed as 
\begin{equation}\label{rep}
y_{rep}=f_{ST5}\left ( X;\theta_{E}  \right ),
\end{equation}
where $f_{ST5}$ signifies the Sentence-T5 model that has been pre-trained. 
The newly-proposed OFC loss is utilized for calculate the loss.
The objective of network pre-training is to obtain a discriminative text representation from Sentence-T5 over the training data by minimizing the loss function, as shown by
\begin{equation}\label{eq_minloss}
\theta _{ST5} = \mathop{\arg \min}\limits_{\theta _{ST5}} \mathcal{L}_{OFC}\left ( X,\hat{y};\theta _{ST5} \right ).
\end{equation}
Algorithm \ref{dmtc} outlines the procedure of the DMTC algorithm.

\begin{breakablealgorithm}
    \small
    \caption{DMTC algorithm for multi-label text classification}
    \label{dmtc}
    \begin{algorithmic}[1]
    \Input Class labels $L=\left \{ l_{1},l_{2},\cdots ,l_{m} \right \}$ and test instance $x$
    \Output Probability distribution $y$ of $x$
        \State $X \gets \left \{  \right \}$
        \For{$l_{i}$ in $L$}
            \State Construct prompt templates $T_{l_{i}}$ based on $l_{i}$;
            \State $x_{l_{i}} \gets LLaMA\left (T_{l_{i}}  \right )$;
            \State Put $x_{l_{i}}$ into $X$;
        \EndFor
        \State $\theta _{E} \gets$ Load the pre-trained Sentence-T5 model;
        \State $\theta _{ST5} \gets$ Pre-train Sentence-T5 to minimize $\mathcal{L}_{OFC}$ (Eq. (\ref{eq21}));
        \State $\left ( \theta_{ST5},\theta_{FC} \right ) \gets$ Fine-tune Sentence-T5 to minimize overall loss function (cross-entropy loss function $\mathcal{L}_{CE}$);
        \State $y \gets f_{Sigmoid}\left ( f_{FC}\left ( f_{ST5}\left ( x;\theta _{ST5} \right );\theta _{FC} \right ) \right )$;
        \State \Return $y$
    \end{algorithmic}
\end{breakablealgorithm}

\section{EXPERIMENTS AND EVALUATION}
\label{experiments}
\subsection{Dataset}
Since DMTC is "data‑less", no external training data is required. Instead, a test set of 918 manually expert-curated samples was established, covering eight distinct AI model categories in maritime transportation. Table \ref{dataset_1} presents an example query for each single scenario, while Table \ref{dataset_2} provides examples of multi-labelled scenarios.

\begin{table*}[htbp]
\small
\centering
\caption{Example for each single class (involve only one maritime AI model) in our constructed data set}
\label{dataset_1}
\begin{tabular}{|p{6.0cm}|p{9cm}|}
\hline
\textbf{Scenarios} & \textbf{Query example} \\
\hline
Long-range ETA in maritime & Forecast the ETA to next port of vessel with MMSI 564765123 \\
\hline
Arrival time to pilotage boarding ground & what is the arrival time of vessel MOUNT ST to the Eastern Boarding Ground ``B" (PEBGB) of Singapore?\\
\hline
Direct berthing to port & Vessel name: Achéron; MMSI: 409011089, chance to get a direct berth when she arrives at the port\\
\hline
Vehicular waiting time after arrival & Tanker vessel ACTIVE (IMO: 6578109, MMSI: 2877619019, call sign: TYRA). How long time will she wait for a berth after arrival?\\
\hline
Ship fuel consumption & How much LSFO is consumed by the tanker vessel for 10 hours?\\
\hline
Berth staying & estimate the time to unberth (ETU) of the container ship NYK CONSTELLATION (IMO: 9337626)\\
\hline
Maritime risk evaluation & Does there any cases about piracy/terrorism detected for the voyage from the Port of Ningbo to Singapore?\\
\hline
Vessel trajectory & Predict the route for container ship OOCL POLAND (IMO 9622588)\\
\hline
\end{tabular}
\end{table*}

\begin{table*}[htbp]
\small
\centering
\caption{Example for some multi-label classes (involve multiple maritime AI models) in our constructed data set}
\label{dataset_2}
\begin{tabular}{|p{5.0cm}|p{10.0cm}|}
\hline
\textbf{Scenarios} & \textbf{Query example} \\
\hline
Direct berthing to port & \multirow{2}{=}{How big is the change to get a direct berth for the vessel? If not, how long time will vessel Aeternum Dread wait at the anchorage?} \\
\cline{1-1}
Vehicular waiting time after arrival & \\
\hline

Long-range ETA in maritime & \multirow{3}{=}{How many hours left for the vessel Ore Italia to arrive at the destination, and the direct and indirect berthing rate? How much LSFO can be utilized during the past 3 hours?} \\
\cline{1-1}
Direct berthing to port & \\
\cline{1-1}
Ship fuel consumption & \\
\hline
\end{tabular}
\end{table*}

\subsection{Results and analyses}
Table \ref{table_comp_1}  summarizes the performance of the proposed pipeline (DMTC) in multi-label intention classification task: 70.15\% in accuracy, 95.92\% in AUC, and 5.35\% of Hamming loss. 
A comprehensive benchmark against popular baselines reveals clear trends. Two dense embeddings (GloVe, Word2Vec) and two contextual embeddings (BERT, MPNet) were paired with conventional machine learning classifiers (NB, SVM) and neural network classifiers (MLP, LSTM, CNN, Bi‑LSTM). Models using BERT or MPNet consistently outperform those using GloVe and Word2Vec: all BERT/MPNet variants exceed 60\% accuracy, 93\% AUC, and 81\% F1‑score, whereas GloVe/Word2Vec models cap at 38\% accuracy, 89\% AUC, and 64\% F1‑score. Among dense‑embedding classifiers, SVM and Bi‑LSTM lead; with contextual embeddings, MLP, LSTM, and CNN perform similarly, yielding 60.02–62.96\% accuracy, 93.80–95.79\% AUC, and 81.21–83.58\% F1‑score. This disparity underscores the inadequacy of traditional embeddings for fine‑grained multiclass intent classification.

\begin{table*}[htbp]
\footnotesize
\centering
\caption{Performance comparison between different machine learning methods}
\label{table_comp_1}
\begin{tabular}{|p{3.1cm}|p{1.4cm}|p{1.4cm}|p{1.9cm}|p{1.4cm}|p{1.4cm}|p{1.1cm}|p{0.9cm}|p{0.9cm}|}
\hline
Model & Accuracy (\%) & Hamming Loss (\%) & Jaccard Similarity (\%) & F1-Score (\%) & Precision (\%) & Recall (\%) & MCC (\%) & AUC (\%) \\
\hline
GloVe + NB & 15.47 & 21.83 & 25.82 & 41.04 & 43.02 & 39.24 & 28.00 & 75.64 \\
\hline
GloVe + SVM & 31.70 & 14.39 & 42.64 & 59.76 & 65.15 & 55.20 & 52.37 & 84.72 \\
\hline
GloVe + CNN & 23.53 & 16.26 & 32.20 & 48.71 & 62.58 & 39.87 & 41.74 & 76.48 \\
\hline
GloVe + Bi-LSTM & 37.91 & 12.91 & 45.33 & 62.38 & 71.58 & 55.27 & 56.18 & 83.42 \\
\hline
Word2Vec + NB & 13.40 & 21.96 & 25.60 & 40.76 & 42.66 & 39.03 & 27.62 & 74.98 \\
\hline
Word2Vec + SVM & 37.04 & 12.39 & 46.94 & 63.89 & 73.32 & 56.61 & 56.91 & 88.75 \\
\hline
Word2Vec + CNN & 35.29 & 13.73 & 44.22 & 61.32 & 67.48 & 56.19 & 56.00 & 86.04 \\
\hline
Word2Vec + Bi-LSTM & 36.17 & 12.06 & 46.24 & 63.24 & 77.13 & 53.59 & 57.89 & 84.50 \\
\hline
BERT + CNN & 60.02 & 7.41 & 68.70 & 81.45 & 79.07 & 83.97 & 77.81 & 94.70 \\
\hline
BERT + LSTM & 61.44 & 6.69 & 71.80 & 83.58 & 79.76 & \textbf{87.90} & 80.48 & 95.79 \\
\hline
MPNet + CNN & 62.96 & 6.97 & 69.86 & 82.26 & 81.08 & 83.47 & 78.56 & 94.19 \\
\hline
MPNet + MLP & 62.31 & 7.12 & 69.45 & 81.97 & 80.39 & 83.61 & 78.31 & 94.04 \\
\hline
MPNet + LSTM & 61.98 & 7.43 & 68.37 & 81.21 & 79.51 & 82.98 & 77.63 & 93.80 \\
\hline
\textbf{DMTC} & \textbf{70.15} & \textbf{5.35} & \textbf{75.53} & \textbf{86.06} & \textbf{86.83} & 85.30 & \textbf{83.32} & \textbf{95.92} \\
\hline
\end{tabular}
\end{table*}

\subsection{Ablation studies}
Table \ref{table_comp_2} compares the introduced Sentence-T5 with other popular LLM-based contextual embedding methods, including MiniLM, RoBERTa, BERT, and MPNet.
Overall, Sentence-T5 outperforms  other LLMs in text representation. Among the baselines, Sentence‑T5 leads across all metrics, with MPNet a close second (66.88\% accuracy, 83.66\% F1, 93.92\% AUC). BERT and RoBERTa deliver moderate performance, while MiniLM trails, underscoring the importance of high‑quality contextual embeddings for fine‑grained intent tasks. 

\begin{table*}[htbp]
\footnotesize
\centering
\caption{Performance comparison between different embedding models}
\label{table_comp_2}
\begin{tabular}{|p{3.1cm}|p{1.4cm}|p{1.4cm}|p{1.9cm}|p{1.4cm}|p{1.4cm}|p{1.1cm}|p{0.9cm}|p{0.9cm}|}
\hline
Model & Accuracy (\%) & Hamming Loss (\%) & Jaccard Similarity (\%) & F1-Score (\%) & Precision (\%) & Recall (\%) & MCC (\%) & AUC (\%) \\
\hline
wth MiniLM & 54.47 & 7.90 & 63.52 & 77.69 & 85.74 & 71.03 & 74.21 & 93.97 \\
\hline
wth RoBERTa & 60.78 & 7.67 & 66.65 & 79.99 & 80.88 & 79.11 & 76.17 & 89.91 \\
\hline
wth BERT & 60.89 & 7.16 & 69.82 & 82.23 & 79.13 & \textbf{85.58} & 78.44 & 94.04 \\
\hline
wth MPNet & 66.88 & 6.29 & 71.91 & 83.66 & 84.14 & 83.19 & 80.54 & 93.92 \\
\hline
\textbf{wth Sentence-T5} & \textbf{70.15} & \textbf{5.35} & \textbf{75.53} & \textbf{86.06} & \textbf{86.83} & 85.30 & \textbf{83.32} & \textbf{95.92} \\
\hline
\end{tabular}
\end{table*}

Table \ref{table_comp_3} evaluates the impact of various contrastive losses on classification accuracy.  The proposed OFC loss achieves 75.15\% accuracy—at least 0.98\% higher than angular‑margin (AMC), N‑pair (NP), cosine‑similarity (CS), and online‑contrastive (OC) losses. The results show that AMC loss and NP loss are not well-suited for this task. This is probably because AMC loss is particularly beneficial tasks where embeddings require clear class separation along with a specific angular relationship within the same class. NP loss is designed for tasks involving multi-label classification, which may not be suitable for this multi-label classification problem. CS and OC offer stronger baselines, with OC benefiting from hard‑sample mining, yet both fall short of the OFC’s emphasis on challenging examples and label combination complexity. 

\begin{table*}[htbp]
\footnotesize
\centering
\caption{Performance comparison between different contrastive loss functions}
\label{table_comp_3}
\begin{tabular}{|p{3.1cm}|p{1.4cm}|p{1.4cm}|p{1.9cm}|p{1.4cm}|p{1.4cm}|p{1.1cm}|p{0.9cm}|p{0.9cm}|}
\hline
Model & Accuracy (\%) & Hamming Loss (\%) & Jaccard Similarity (\%) & F1-Score (\%) & Precision (\%) & Recall (\%) & MCC (\%) & AUC (\%) \\
\hline
wth AMC loss & 45.64 & 9.83 & 56.03 & 71.82 & 80.70 & 64.70 & 66.92 & 91.04 \\
\hline
wth NP loss & 52.61 & 8.27 & 61.80 & 76.39 & 85.47 & 69.06 & 72.51 & 92.59 \\
\hline
wth CS loss & 67.54 & 5.69 & 73.64 & 84.82 & 87.69 & 82.14 & 81.88 & 93.28 \\
\hline
wth OC loss & 69.17 & 5.53 & 74.89 & 85.64 & 86.13 & 85.16 & 82.63 & 95.69 \\
\hline
\textbf{wth OFC loss} & \textbf{70.15} & \textbf{5.35} & \textbf{75.53} & \textbf{86.06} & \textbf{86.83} & \textbf{85.30} & \textbf{83.32} & \textbf{95.92} \\
\hline
\end{tabular}
\end{table*}

Table \ref{table_comp_4} examines the effect of incorporating contrastive learning into the overall DMTC framework. Adding the OFC loss yields consistent gains: accuracy jumps from 68.08\% to 70.15\% (+2.07\%), F1‑score climbs from 84.58\% to 86.06\% (+1.48\%), and Hamming loss drops from 5.80\% to 5.35\% (–0.45\%). All eight evaluation metrics improve, confirming that contrastive objectives bolster multi‑label classification robustness.

\begin{table*}[htbp]
\footnotesize
\centering
\caption{Effects of using the contrastive learning paradigm}
\label{table_comp_4}
\begin{tabular}{|p{3.6cm}|p{1.4cm}|p{1.4cm}|p{1.9cm}|p{1.4cm}|p{1.4cm}|p{1.1cm}|p{0.9cm}|p{0.9cm}|}
\hline
Model & Accuracy (\%) & Hamming Loss (\%) & Jaccard Similarity (\%) & F1-Score (\%) & Precision (\%) & Recall (\%) & MCC (\%) & AUC (\%) \\
\hline
w/o contrastive learning & 68.08 & 5.80 & 73.27 & 84.58 & 87.16 & 82.14 & 81.26 & 95.65 \\
\hline
\textbf{wth contrastive learning} & \textbf{70.15} & \textbf{5.35} & \textbf{75.53} & \textbf{86.06} & \textbf{86.83} & \textbf{85.30} & \textbf{83.32} & \textbf{95.92} \\
\hline
\end{tabular}
\end{table*}

Table \ref{table_comp_5} benchmarks DMTC against end‑to‑end LLM pipelines (GPT‑4, GPT‑4o) under identical prompts and test data. DMTC achieves a subset accuracy of 70.15\% and Jaccard similarity of 75.53\%, far exceeding GPT‑4’s 32.14\%, and GPT‑4o’s 29.30\%. These results highlight the limitations of prompt‑only LLM approaches in specialized, multi‑label intent classification and validate the effectiveness of the DMTC design.

\begin{table*}[htbp]
\footnotesize
\centering
\caption{Performance comparison between different LLMs}
\label{table_comp_5}
\begin{threeparttable}
\begin{tabular}{|p{2.0cm}|p{1.3cm}|p{1.4cm}|p{2.0cm}|p{1.3cm}|p{1.2cm}|p{1cm}|p{0.8cm}|}
\hline
Model & Accuracy (\%) & Hamming Loss (\%) & Jaccard Similarity (\%) & F1-Score (\%) & Precision (\%) & Recall (\%) & MCC (\%) \\
\hline
GPT-4 & 32.14 & 16.00 & 40.51 & 57.66 & 59.13 & 56.26 & 53.05 \\
\hline
GPT-4o & 29.30 & 16.37 & 40.99 & 58.15 & 57.59 & 58.72 & 53.22 \\
\hline
\textbf{DMTC} & \textbf{70.15} & \textbf{5.35} & \textbf{75.53} & \textbf{86.06} & \textbf{86.83} & \textbf{85.30} & \textbf{83.32}\\
\hline
\end{tabular}
\textit{Note:} Since LLMs do not generate class probability distributions, measuring the AUC is not feasible.
\end{threeparttable}
\end{table*}

\section{Conclusions}
\label{conclusion}
This work has presented \textbf{DMTC}, a modular, data‑free pipeline for multi‑label intention classification in transportation applications. The key components includes: 1) \textbf{Zero‑shot synthetic data generation} via prompt‑engineered large language models, removing the reliance on costly, manually labeled corpora; 2)  \textbf{Sentence‑T5 semantic embeddings}, which capture nuanced contextual information in user queries and enable robust multi‑class discrimination; 3) \textbf{Online focal‑contrastive (OFC) loss}, a novel contrastive objective that focuses on hard sample pairs to sharpen inter‑class separability and improve generalization.
Extensive experiments demonstrate that DMTC achieves state‑of‑the‑art performance—exceeding 86 \% F1‑score, 95\% AUC, and reducing Hamming loss below 5.5\%—while outperforming both classical and LLM‑based baselines. These results validate the effectiveness of data‑free learning and contrastive feature learning for fine‑grained intent understanding in complex, multi‑label scenarios.
Looking ahead, DMTC’s plug‑and‑play design can be extended beyond maritime use cases to broader transportation domains, including logistics, traffic management, and autonomous mobility. Future work will explore end‑to‑end integration with downstream agentic AI modules, dynamic prompt adaptation for evolving intent taxonomies, and unsupervised refinement of embedding spaces to further reduce human intervention and enhance scalability.

\section{ACKNOWLEDGEMENT}
This study is supported under Maritime AI Research Programme (Grant number SMI-2022-MTP-06 funded by Singapore Maritime Institute).

\bibliography{mybibfile}

@article{xiao2022next,
  title={Next-generation vessel traffic services systems—From “passive” to “proactive”},
  author={Xiao, Zhe and Fu, Xiuju and Zhao, Liangbin and Zhang, Liye and Teo, Tze Kern and Li, Ning and Zhang, Wanbing and Qin, Zheng},
  journal={IEEE Intell. Transp. Syst. Mag.},
  volume={15},
  number={1},
  pages={363--377},
  year={2022},
  publisher={IEEE}
}

@article{zhang2022vessel,
  title={Vessel trajectory prediction in maritime transportation: Current approaches and beyond},
  author={Zhang, Xiaocai and Fu, Xiuju and Xiao, Zhe and Xu, Haiyan and Qin, Zheng},
  journal={IEEE Trans. Intell. Transp. Syst.},
  year={2022},
  publisher={IEEE}
}

@article{liu2023multi,
  title={A multi-task deep learning model integrating ship trajectory and collision risk prediction},
  author={Liu, Tao and Xu, Xiang and Lei, Zhengling and Zhang, Xiaocai and Sha, Mei and Wang, Fang},
  journal={Ocean Eng.},
  volume={287},
  pages={115870},
  year={2023},
  publisher={Elsevier}
}

@article{zhang2024viewpoint,
  title={A Viewpoint Adaptation Ensemble Contrastive Learning framework for vessel type recognition with limited data},
  author={Zhang, Xiaocai and Xiao, Zhe and Fu, Xiuju and Wei, Xiaoyang and Liu, Tao and Yan, Ran and Qin, Zheng and Zhang, Jianjia},
  journal={Expert Syst. Appl.},
  volume={238},
  pages={122191},
  year={2024},
  publisher={Elsevier}
}

@article{chen2023prediction,
  title={Prediction of harbour vessel fuel consumption based on machine learning approach},
  author={Chen, Zhong Shuo and Lam, Jasmine Siu Lee and Xiao, Zengqi},
  journal={Ocean Eng.},
  volume={278},
  pages={114483},
  year={2023},
  publisher={Elsevier}
}

@inproceedings{maghrebi2015complementing,
  title={Complementing travel diary surveys with twitter data: application of text mining techniques on activity location, type and time},
  author={Maghrebi, Mojtaba and Abbasi, Alireza and Rashidi, Taha Hossein and Waller, S Travis},
  booktitle={2015 IEEE 18th International Conference on Intelligent Transportation Systems},
  pages={208--213},
  year={2015},
  organization={IEEE}
}

@article{ali2019fuzzy,
  title={Fuzzy ontology and LSTM-based text mining: a transportation network monitoring system for assisting travel},
  author={Ali, Farman and El-Sappagh, Shaker and Kwak, Daehan},
  journal={Sensors},
  volume={19},
  number={2},
  pages={234},
  year={2019},
  publisher={MDPI}
}

@inproceedings{styawati2022sentiment,
  title={Sentiment analysis on online transportation reviews using Word2Vec text embedding model feature extraction and support vector machine (SVM) algorithm},
  author={Styawati, Styawati and Nurkholis, Andi and Aldino, Ahmad Ari and Samsugi, Selamet and Suryati, Emi and Cahyono, Ryan Puji},
  booktitle={2021 International Seminar on Machine Learning, Optimization, and Data Science (ISMODE)},
  pages={163--167},
  year={2022},
  organization={IEEE}
}

@inproceedings{klaithin2016traffic,
  title={Traffic information extraction and classification from Thai Twitter},
  author={Klaithin, Supon and Haruechaiyasak, Choochart},
  booktitle={2016 13th International Joint Conference on Computer Science and Software Engineering (JCSSE)},
  pages={1--6},
  year={2016},
  organization={IEEE}
}

@article{chen2018detecting,
  title={Detecting traffic information from social media texts with deep learning approaches},
  author={Chen, Yuanyuan and Lv, Yisheng and Wang, Xiao and Li, Lingxi and Wang, Fei-Yue},
  journal={IEEE Trans. Intell. Transp. Syst.},
  volume={20},
  number={8},
  pages={3049--3058},
  year={2018},
  publisher={IEEE}
}

@inproceedings{chen2017convolutional,
  title={A convolutional neural network for traffic information sensing from social media text},
  author={Chen, Yuanyuan and Lv, Yisheng and Wang, Xiao and Wang, Fei-Yue},
  booktitle={2017 IEEE 20th International Conference on Intelligent Transportation Systems (ITSC)},
  pages={1--6},
  year={2017},
  organization={IEEE}
}

@article{azhar2023detection,
  title={Detection and prediction of traffic accidents using deep learning techniques},
  author={Azhar, Anique and Rubab, Saddaf and Khan, Malik M and Bangash, Yawar Abbas and Alshehri, Mohammad Dahman and Illahi, Fizza and Bashir, Ali Kashif},
  journal={Cluster Comput.},
  volume={26},
  number={1},
  pages={477--493},
  year={2023},
  publisher={Springer}
}

@article{wan2020empowering,
  title={Empowering real-time traffic reporting systems with nlp-processed social media data},
  author={Wan, Xiangpeng and Lucic, Michael C and Ghazzai, Hakim and Massoud, Yehia},
  journal={IEEE Open J. Intell. Transp. Syst.},
  volume={1},
  pages={159--175},
  year={2020},
  publisher={IEEE}
}

@article{babbar2023real,
  title={Real-time traffic, accident, and potholes detection by deep learning techniques: a modern approach for traffic management},
  author={Babbar, Sarthak and Bedi, Jatin},
  journal={Neural Comput. Appl.},
  volume={35},
  number={26},
  pages={19465--19479},
  year={2023},
  publisher={Springer}
}

@inproceedings{jidkov2020enabling,
  title={Enabling maritime risk assessment using natural language processing-based Deep Learning Techniques},
  author={Jidkov, Vladislav and Abielmona, Rami and Teske, Alexander and Petriu, Emil},
  booktitle={2020 IEEE Symposium Series on Computational Intelligence (SSCI)},
  pages={2469--2476},
  year={2020},
  organization={IEEE}
}

@article{khodadadi2022natural,
  title={A natural language processing and deep learning based model for automated vehicle diagnostics using free-text customer service reports},
  author={Khodadadi, Ali and Ghandiparsi, Soroush and Chuah, Chen-Nee},
  journal={Mach. Learn. Appl.},
  volume={10},
  pages={100424},
  year={2022},
  publisher={Elsevier}
}

@inproceedings{teske2018automatic,
  title={Automatic identification of maritime incidents from unstructured articles},
  author={Teske, Alexander and Falcon, Rafael and Abielmona, Rami and Petriu, Emil},
  booktitle={2018 IEEE Conference on Cognitive and Computational Aspects of Situation Management (CogSIMA)},
  pages={42--48},
  year={2018},
  organization={IEEE}
}

@article{suat2022extraction,
  title={Extraction and analysis of social networks data to detect traffic accidents},
  author={Suat-Rojas, Nestor and Gutierrez-Osorio, Camilo and Pedraza, Cesar},
  journal={Inf.},
  volume={13},
  number={1},
  pages={26},
  year={2022},
  publisher={MDPI}
}

@inproceedings{neruda2021traffic,
  title={Traffic event detection from Twitter using a combination of CNN and BERT},
  author={Neruda, Gregorius Aria and Winarko, Edi},
  booktitle={2021 International Conference on Advanced Computer Science and Information Systems (ICACSIS)},
  pages={1--7},
  year={2021},
  organization={IEEE}
}

@article{osorio2021social,
  title={Social media semantic perceptions on Madrid Metro system: Using Twitter data to link complaints to space},
  author={Osorio-Arjona, Joaqu{\'\i}n and Horak, Jiri and Svoboda, Radek and Garc{\'\i}a-Ru{\'\i}z, Yolanda},
  journal={Sustainable Cities Soc.},
  volume={64},
  pages={102530},
  year={2021},
  publisher={Elsevier}
}

@inproceedings{zhang2023prediction,
  title={Prediction of vessel arrival time to pilotage area using multi-data fusion and deep learning},
  author={Zhang, Xiaocai and Fu, Xiuju and Xu, Haiyan and Wei, Xiaoyang and Koh, Jimmy and Ogawa, Daichi and Qin, Zheng},
  booktitle={2023 IEEE 26th International Conference on Intelligent Transportation Systems (ITSC)},
  year={2023},
  organization={IEEE}
}

@inproceedings{raksachat2023improving,
  title={Improving a text classifier using text augmentation: road traffic content from Twitter},
  author={Raksachat, Thawatchai and Chawuthai, Rathachai},
  booktitle={2023 20th International Conference on Electrical Engineering/Electronics, Computer, Telecommunications and Information Technology},
  pages={1--4},
  year={2023},
  organization={IEEE}
}

@article{pan2023ernie,
  title={Ernie-Gram BiGRU Attention: An Improved Multi-Intention Recognition Model for Air Traffic Control},
  author={Pan, Weijun and Jiang, Peiyuan and Wang, Zhuang and Li, Yukun and Liao, Zhenlong},
  journal={Aerosp.},
  volume={10},
  number={4},
  pages={349},
  year={2023},
  publisher={MDPI}
}

@article{zuluaga2022atco2,
  title={ATCO2 corpus: A Large-Scale Dataset for Research on Automatic Speech Recognition and Natural Language Understanding of Air Traffic Control Communications},
  author={Zuluaga-Gomez, Juan and Vesel{\`y}, Karel and Sz{\"o}ke, Igor and Motlicek, Petr and Kocour, Martin and Rigault, Mickael and Choukri, Khalid and Prasad, Amrutha and Sarfjoo, Seyyed Saeed and Nigmatulina, Iuliia and others},
  journal={arXiv preprint arXiv:2211.04054},
  year={2022}
}

@article{ni2021sentence,
  title={Sentence-t5: Scalable sentence encoders from pre-trained text-to-text models},
  author={Ni, Jianmo and {\'A}brego, Gustavo Hern{\'a}ndez and Constant, Noah and Ma, Ji and Hall, Keith B and Cer, Daniel and Yang, Yinfei},
  journal={arXiv preprint arXiv:2108.08877},
  year={2021}
}

@inproceedings{zhang2024maritime,
  title={Maritime-Context Text Identification for Connecting Artificial Intelligence (AI) Models},
  author={Zhang, Xiaocai and Lim, Hur and Fu, Xiuju and Wang, Ke and Xiao, Zhe and Qin, Zheng},
  booktitle={2024 IEEE Conference on Artificial Intelligence (CAI)},
  pages={899--904},
  year={2024},
  organization={IEEE}
}

@article{zhang2024dynamic,
  title={A Dynamic Context-Aware Approach for Vessel Trajectory Prediction Based on Multi-Stage Deep Learning},
  author={Zhang, Xiaocai and Fu, Xiuju and Xiao, Zhe and Xu, Haiyan and Zhang, Wanbing and Koh, Jimmy and Qin, Zheng},
  journal={IEEE Trans. Intell. Veh.},
  year={2024},
  publisher={IEEE}
}

@article{lee2024analysis,
  title={Analysis of Bi-LSTM CRF Series Models for Semantic Classification of NAVTEX Navigational Safety Messages},
  author={Lee, Changui and Cho, Hoyeon and Lee, Seojeong},
  journal={J. Mar. Sci. Eng.},
  volume={12},
  number={9},
  pages={1518},
  year={2024},
  publisher={MDPI}
}

@article{pullanikkat2024utilizing,
  title={Utilizing the Twitter social media to identify transportation-related grievances in Indian cities},
  author={Pullanikkat, Rahul and Poddar, Soham and Das, Anik and Jaiswal, Tushar and Singh, Vivek Kumar and Basu, Moumita and Ghosh, Saptarshi},
  journal={Social Network Anal. Min.},
  volume={14},
  number={1},
  pages={118},
  year={2024},
  publisher={Springer}
}

@inproceedings{hui2023atcbert,
  title={ATCBERT: Few-shot Intent Recognition for Air Traffic Control Instruction Understanding},
  author={Hui, Yi and Cai, Kaiquan and Zhang, Minghua and Qian, Shengsheng and Yang, Yang},
  booktitle={2023 IEEE 26th International Conference on Intelligent Transportation Systems (ITSC)},
  pages={4510--4517},
  year={2023},
  organization={IEEE}
}

@article{zhang2025natural,
  title={Natural language processing and text mining in transportation: Current status, challenges, and future roadmap},
  author={Zhang, Xiaocai and Gao, Ruobin and Xiao, Zhe and Wang, Ke and Liu, Tao and Liang, Maohan and Zhang, Jianjia},
  journal={Expert Syst. Appl.},
  pages={129050},
  year={2025},
  publisher={Elsevier}
}

\end{document}